\documentclass[10pt,a4paper]{article}

\usepackage[final]{lrec2026} 

\usepackage{times}
\usepackage{latexsym}
\usepackage{hyperref}
\usepackage{graphicx}
\usepackage{amsmath}
\usepackage{natbib}
\usepackage[utf8]{inputenc}
\usepackage{xcolor}
\usepackage{xurl}
\usepackage{booktabs}
\usepackage{newfloat}
\usepackage{listings}
\usepackage{caption}
\usepackage[export]{adjustbox}
\usepackage{lipsum}
\usepackage{multirow}
\usepackage{longtable}

\title{\textsc{Ayn}: A Tiny yet Competitive Indian Legal Language Model Pretrained from Scratch} 

\name{Mitodru Niyogi$^{1}$, Eric Gaussier$^{1}$, Arnab Bhattacharya$^{2}$}

\address{
$^{1}$ Université Grenoble Alpes, CNRS, Grenoble INP, LIG \\
38000 Grenoble, France \\
$^{2}$ Dept. of Computer Science and Engineering, Indian Institute of Technology Kanpur \\
Kanpur, India \\
\texttt{mitodru.niyogi@cnrs.fr, eric.gaussier@imag.fr, arnabb@cse.iitk.ac.in}
}

\abstract{
Decoder-only Large Language Models (LLMs) are currently the model of choice for many Natural Language Processing (NLP) applications. Through instruction fine-tuning and prompting approaches, such LLMs have been efficiently used to solve both general and domain-specific tasks. However, they are costly to train and, to a certain extent, costly to use as well, and one can wonder whether LLMs can be replaced by domain-specific Tiny Language Models (TLMs), which typically contain less than 100M parameters. We address this question in this study by comparing the performance of an 88M TLM pretrained from scratch for 185 A100 hours on a specific domain with a domain-specific tokenizer (here, the Indian legal domain) with LLMs of various sizes between 1B and 8B for solving domain-specific tasks.  
We show in particular that our legal TLM, \textsc{Ayn}, can indeed outperform LLMs up to 80 times larger on the legal case judgment prediction task, rival LLMs up to 30 times larger on the summarization task, and still be competitive with these larger LLMs on general tasks.
\\ \newline 
}

\newcommand{\ab}[1]{\textcolor{red}{AB: #1}}
\newcommand{\mn}[1]{\textcolor{blue}{Mitodru: #1}}

\newcommand{\comment}[1]{}

\newcommand{\cutabove}{\vspace*{-2mm}}
\newcommand{\cutspace}{\vspace*{-5mm}}

\begin{document}

\maketitleabstract

\section{Introduction}

Decoder-only Large Language Models (LLMs) pretrained on large amounts of diverse data have been used to solve both general and domain-specific tasks.  
Different approaches have been proposed to adapt or use LLMs in specific domains, such as distillation \citep{yao-etal-2021-adapt,tan-etal-2023-gkd,wen-etal-2023-f,gao-etal-2024-kd}, continual pretraining and supervised fine-tuning \citep{gururangan-etal-2020-dont,colombo2024saullm7bpioneeringlargelanguage,li2024bladeenhancingblackboxlarge,chen2023meditron70b}, retrieval augmentation \citep{cheng2024adapting, sachidananda-etal-2021-efficient}, and zero-shot prompting \cite{10806871,Li2023PromptingLL,Fahes2022PDAPZ}. 
However, since acquiring large-scale corpora and performing training (and, to a lesser extent, inference) are costly processes, one may wonder whether it is possible to rely on tiny language models (under 100 million parameters) for solving domain-specific (e.g., legal) tasks, pretrained from scratch on limited domain data\footnote{having a corpus with less than 150 million words}, thereby avoiding the need to first train a large model and subsequently distill it. This cost challenge is even more acute in the legal domain, where annotated data is scarce, expensive, and time-consuming to produce \cite{chalkidis-etal-2020-legal,zheng2021does}. 

Legal language processing poses unique challenges: specialized and archaic vocabulary (e.g., ``hereinafter'', ``notwithstanding'', etc.), long and nested sentences, and complex citation structures (e.g., ``Section 3(1)(b) of the Act'') \citep{chalkidis-etal-2019-neural,chalkidis-etal-2020-legal,niklaus-giofre-2023-pretrain}. Indian legal texts further involve multilingual code-switching and domain-specific jargon, complicating tokenization and modeling \citep{ganguly2023legal}. General-purpose tokenizers often fragment these expressions inefficiently, thereby harming model learning and performance. To address this, we pretrain a legal domain-specific tokenizer tailored to Indian legal text that captures multi-word legal expressions, terminology, citations, and legal language variations more effectively. 

Existing LLMs also exhibit Western-centric biases \citep{johnson2022ghostmachineamericanaccent,10.1093/pnasnexus/pgae346,gallegos2024bias}, as they are predominantly trained on datasets originating from Western legal systems \citep{hendersonkrass2022pileoflaw,naous-xu-2025-origin}. 
This bias limits their effectiveness and fairness when applied to other jurisdictions, such as India \cite{DBLP:journals/corr/abs-2309-08573,rinki2025measuringsouthasianbiases,adilazuarda-etal-2024-towards}, which features a distinct legal tradition, multilingualism, and diverse socio-cultural contexts.

Our work tackles this gap by training a tiny (88M) parameter decoder-only model on a curated Indian legal corpus. We evaluate it on legal and general benchmarks, and compare it to generalist LLMs, continually pretrained LLMs, and similarly sized baselines. By focusing on tiny, domain-specialized models, we propose an efficient and accessible alternative, particularly suited for resource-constrained settings, to large-scale model adaptation. Beyond the domain of Indian legal system, our findings contribute to addressing the pervasive bias in existing LLMs and suggest pathways toward equitable and effective legal NLP tools in under-represented domains globally.

More precisely, we aim to answer the following research questions:
How does a decoder-only tiny language model (TLM)\footnote{We consider here a language model to be \textit{tiny} if it has below 100M parameters.} under 100M parameters pretrained from scratch on legal data:
\begin{itemize}
    \item {\bf RQ1:} compare to decoder-only general-purpose LLMs on legal tasks?
    \item {\bf RQ2:} generalize to general LLM benchmark tasks such as multiple-choice question answering and classification?
\end{itemize}

Our main contributions parallel these research questions and are four-fold:
\begin{itemize}
    
    \item[1.] Resource Contributions:
    
    \begin{itemize}
        \item[(i)] We propose a new corpus for Indian Supreme Court case documents by extending the ILDC \citeplanguageresource{malik-etal-2021-ildc} corpus with 3,046 Indian Supreme Court case documents (May 2020 - Dec 2023), the Constitution of India, and the Indian Penal Code, totaling 142.6 million words. 
        \item[(ii)] We develop a legal domain-specific BPE tokenizer from scratch for Indian Supreme Court case documents.
        \item[(iii)] We build a 88M parameter decoder-only TLM pretrained from scratch on the above corpus with a context size of 8192 on a single A100 GPU for 185 hours.
\end{itemize}

    \item[2.] We empirically study the performance of a TLM with fewer than 100M parameters and compare it to larger, well-known LMs with sizes between 1B and 8B parameters on two different legal tasks: one generative and one classification, as well as on general NLP tasks.
 \end{itemize}
 
The remainder of the paper is organized as follows: Section~\ref{sec:related-work} describes the related work, Section~\ref{sec:data} presents the data used for pretraining, and Section~\ref{sec:model} and Section~\ref{sec:training} present the model and training procedure. The evaluation of our model is detailed in Section~\ref{sec:eval} before Section~\ref{sec:conclusion} concludes.

\begin{table}[t]
\centering
\resizebox{\columnwidth}{!}{
\begin{tabular}{p{0.30\columnwidth} l r}
\toprule
\textbf{Pre-training Corpus} & \textbf{Timeline}  & \textbf{\#Words} \\ \midrule
34,816 SCI Cases (ILDC) & 1947 - April 2020 & 111,469,509     \\ 
3,046 SCI Cases (ours)& May 2020 - Dec 2023 &  30,904,905    \\ 
Total: 38,222 SCI Cases & 1947 - 2023 & 142,374,414     \\
\midrule
Constitution of India (ours) & 2024 &  139,956\\
Penal Code of India (ours) & 2024 & 77,633\\
\midrule
\bf Total & 1947 - 2024 & 142,592,003 \\ \bottomrule
\end{tabular}
}
\caption{Pretraining Corpus}
\label{tab:corpus}
\end{table}

\section{Related Work}
\label{sec:related-work}


Large Language Models (LLMs) have predominantly been pretrained on a variety of data sources ranging from web corpora and books to scientific articles, and also on legal data, but they are primarily exposed to Western legal data. However, legal data typically represent only a small fraction of the entire pretraining dataset. A common approach for domain adaptation of LLMs has been either continual pretraining on domain-specific data or domain-specific pretraining from scratch, such as in the biomedical domain \citep{10.1145/3458754,bolton2024biomedlm27bparameterlanguage}, where the authors demonstrated the advantage of domain-specific pretraining from scratch. In those works, they replicate the exact model configurations of BERT \citep{devlin-etal-2019-bert} or GPT-2 \citep{Radford2019LanguageMA} to train a biomedical model from scratch and compare it with general LLMs and continually pretrained LLMs for domain adaptation. 

In our case, the focus is not on replicating an existing model configuration for domain-specific pretraining, but rather on investigating whether domain-specific pretraining from scratch for a tiny model with fewer than 100 million parameters can be competitive for legal domain adaptation as well as for general NLP tasks across classification and multiple-choice question answering. Talking about legal domain adaptation, existing works have traditionally focused on performing additional continual pretraining on legal data and supervised fine-tuning for legal-specific tasks \citep{10.1007/978-3-031-56060-6_15}, or on integrating domain-specific knowledge with external databases to enhance the performance of LLMs for legal tasks. Notable examples include the Chinese legal LlawyerLlama \citep{huang2023lawyer}, a LLaMA model trained on a large-scale legal dataset; ChatLaw \citep{cui2023chatlaw}, a legal LLM with integrated external knowledge bases; InLegalLLaMA \citep{Ghosh2024InLegalLLaMAIL}, a Llama-2 model continually pretrained on 10,000 Indian legal documents via a LoRA \citep{hu2022lora} adapter with further instruction tuning; and SaulLM-7B \citep{colombo2024saullm7bpioneeringlargelanguage}, a Mistral 7B LLM continually pretrained and supervised instruction-tuned on English court documents comprising over 30 billion tokens across the USA, UK, Canada, and Europe.

Talking about the legal evaluation of law-specific LLMs requires benchmarks that focus on legal tasks. Since laws and legal processes vary by country and are country-specific, developing a single global legal LLM evaluation benchmark may not be desirable, in our opinion. LegalBench \citeplanguageresource{guha2023legalbench} is a collaboratively constructed legal reasoning benchmark comprising 162 tasks in the USA law context. However, to the best of our knowledge, such exhaustive legal benchmarks in the Indian context do not exist. \citetlanguageresource{shukla-etal-2022-legal} introduced a benchmark dataset for abstractive case summarization, and \citetlanguageresource{nigam-etal-2024-legal} introduced the largest legal judgment prediction benchmark dataset with explanations for the Indian judiciary. \citetlanguageresource{joshi-etal-2024-il} introduced a multilingual legal benchmark in nine Indian languages by consolidating existing benchmarks for eight different legal NLP tasks.


\section{Data}
\label{sec:data}

\subsection{Pretraining Data}

The pretraining data is an extended version of the ILDC \citep{malik-etal-2021-ildc} dataset which initially contains 34,816 Supreme Court of India cases in English. To these, we added 3,046 publicly available\footnote{https://indiankanoon.org} case documents of the Supreme Court from May 2020 to December 2023.
In addition to the Supreme Court cases, we also considered the Constitution of India (in English) and the Indian Penal Code (in English), which we scraped from public sources. Table~\ref{tab:corpus} displays the statistics for our clean, preprocessed pretraining corpus. As the number of tokens varies depending on the tokenizer, we report the total number of (whitespace-separated) words instead.

\subsection{Data Cleaning and Preprocessing}

Case proceedings are unstructured documents that have different formats and lengths. In addition, there are typos and spelling mistakes (since these are typed during the court hearing), which makes preprocessing challenging. We utilized regular expressions to clean the text by removing unnecessary elements and meta-information, such as the case number, dates, judge names, and other introductory details at the beginning of the documents.
In Supreme Court case proceedings, the decisions are
written towards the end of the document. These end
section(s), which directly state the decision, have been
kept here.


For both the Constitution of India and the Indian Penal Code documents, we used the \texttt{tesseract}\footnote{https://github.com/tesseract-ocr}
tool to extract the text from the PDF files and to filter out less relevant information from the first few pages of the PDFs, such as the title page, acknowledgments, table of contents, and other irrelevant information. 

\begin{table*}[t]
\centering
\small
\setlength{\tabcolsep}{4pt}
\begin{tabular}{p{\textwidth}}
\toprule
\textbf{Legal Text} \\
\midrule
Notwithstanding anything contained herein, section 279(1A) of the Act operates as a statutory bar to prosecution under sections 276C and 277 only where the Commissioner, hereinafter referred to as the Authority, has in his discretionary jurisdiction reduced or waived the penalty imposable under section 271(1)(iii) pursuant to section 273A. The mere possibility of such discretionary relief does not invalidate pending criminal proceedings instituted heretofore or actions taken thereunder. \\
\midrule
\textbf{Tokenized by LLaMA-2 32k} \\
\midrule
\texttt{[`\_Not', `with', `standing', `\_anything', `\_contained', `\_here', `in', `,', `\_section', `\_', `2', `7', `9', `(', `1', `A', `)', `\_of', `\_the', `\_Act', `\_oper', `ates', `\_as', `\_a', `\_stat', `ut', `ory', `\_bar', `\_to', `\_pro', `sec', `ution', `\_under', `\_sections', `\_', `2', `7', `6', `C', `\_and', `\_', `2', `7', `7', `\_only', `\_where', `\_the', `\_Commission', `er', `,', `\_here', `ina', `fter', `\_referred', `\_to', `\_as', `\_the', `\_Author', `ity', `,', `\_has', `\_in', `\_his', `\_dis', `cret', `ion', `ary', `\_juris', `diction', `\_reduced', `\_or', `\_wa', `ived', `\_the', `\_penalty', `\_impos', `able', `\_under', `\_section', `\_', `2', `7', `1', `(', `1', `)(', `iii', `)', `\_purs', `u', `ant', `\_to', `\_section', `\_', `2', `7', `3', `A', `.', `\_The', `\_mere', `\_possibility', `\_of', `\_such', `\_dis', `cret', `ion', `ary', `\_relief', `\_does', `\_not', `\_invalid', `ate', `\_pending', `\_criminal', `\_proceed', `ings', `\_instit', `uted', `\_her', `et', `of', `ore', `\_or', `\_actions', `\_taken', `\_there', `under', `.']}
 \\
\midrule
\textbf{Tokenized by our Legal BPE} \\
\midrule
\texttt{[`\_Not', `withstanding', `\_anything', `\_contained', `\_herein', `,', `\_section', `\_', `2', `7', `9', `(', `1', `A', `)', `\_of', `\_the', `\_Act', `\_oper', `ates', `\_as', `\_a', `\_statutory', `\_bar', `\_to', `\_prosecution', `\_under', `\_sections', `\_', `2', `7', `6', `C', `\_and', `\_', `2', `7', `7', `\_only', `\_where', `\_the', `\_Commissioner', `,', `\_herein', `after', `\_referred', `\_to', `\_as', `\_the', `\_A', `uthor', `ity', `,', `\_has', `\_in', `\_his', `\_discretion', `ary', `\_jurisdiction', `\_reduced', `\_or', `\_w', `a', `ived', `\_the', `\_penalty', `\_impos', `able', `\_under', `\_section', `\_', `2', `7', `1', `(', `1', `)(', `iii', `)', `\_pursu', `ant', `\_to', `\_section', `\_', `2', `7', `3', `A', `.', `\_The', `\_mere', `\_poss', `ibility', `\_of', `\_such', `\_discretion', `ary', `\_relief', `\_does', `\_not', `\_invalid', `ate', `\_pending', `\_criminal', `\_proceedings', `\_instituted', `\_here', `to', `fore', `\_or', `\_a', `ctions', `\_taken', `\_thereunder', `.']} \\
\bottomrule
\end{tabular}
\caption{Subword segmentation comparison between LLaMA-2 32k and our Legal-domain BPE tokenizer. Leading \texttt{\_} denotes whitespace token.}
\label{tab:legal_bpe_comparison}
\end{table*}

\section{Model Design}
\label{sec:model}

Our model, \textsc{Ayn}\footnote{Available for download at: \url{https://huggingface.co/collections/mitodru/ayn}}, is based on the transformer \citep{10.5555/3295222.3295349} causal decoder architecture \citep{Radford2019LanguageMA}. Following Chinchilla \citep{NEURIPS2022_c1e2faff} and LLaMa \citep{touvron2023llama}, we used RMSNorm \citep{10.5555/3454287.3455397} with \texttt{norm\_epsilon} = $1\mathrm{e}{-5}$, the SwiGLU \citep{shazeer2020glu} activation function, and an activation hidden size of $\sim\frac{8}{3}d$. Following \citep{JMLR:v24:22-1144}, we removed all biases from dense layers to improve the training stability. We also used weight tying \citep{press-wolf-2017-using} to improve the performance of language model by sharing the weights of the embedding and softmax layers. The hidden dimension of Ayn is 768 with 12 layers and a feedforward layer hidden dimension of 2048. We used the interpolation method of \citep{niyogi2024paramanufamilynovelefficient} for the position indices of tokens in RoPE \citep{10.1016/j.neucom.2023.127063} based scaling with $\theta = 10{,}000$ to train longer sequences in a resource-constrained setting (i.e., a single GPU). Specifically, we applied a shrinking factor obtained by dividing the target context length ($y$) by the permissible context length on a single GPU, while keeping all other hyperparameters constant. This allows the tokens' $position\_ids$ to be adjusted by the shrinking factor. For example, with a permissible context size of 256 on an A100 40G GPU, a target context size of 4096 results in a shrinking factor of 16. A token with $position\_id$ = 4000 becomes 250 ($4000/16$). This enables training with larger context sizes than the available memory would otherwise allow, facilitating the pretraining of generative models with larger context sizes on limited hardware. Applying this scaling technique allowed us to pretrain our legal model with a context size of 8192 on a single GPU. The shrinking factor was set to 32 to achieve a target context size of 8192.

\comment{
The following is the function declaration for our implementation. 
\begin{lstlisting}[language=Python, basicstyle=\small\ttfamily, linewidth=0.9\columnwidth]
query_states, key_states = rotary_pos_emb(
query_states, key_states, cos, sin,
position_ids/shrinking_factor)
\end{lstlisting}
\ab{why suddenly this code?}
\mn{to provide a pseudocode of the method declaration}

}


\begin{figure}[t]
    \centering
    \includegraphics[width=\linewidth]{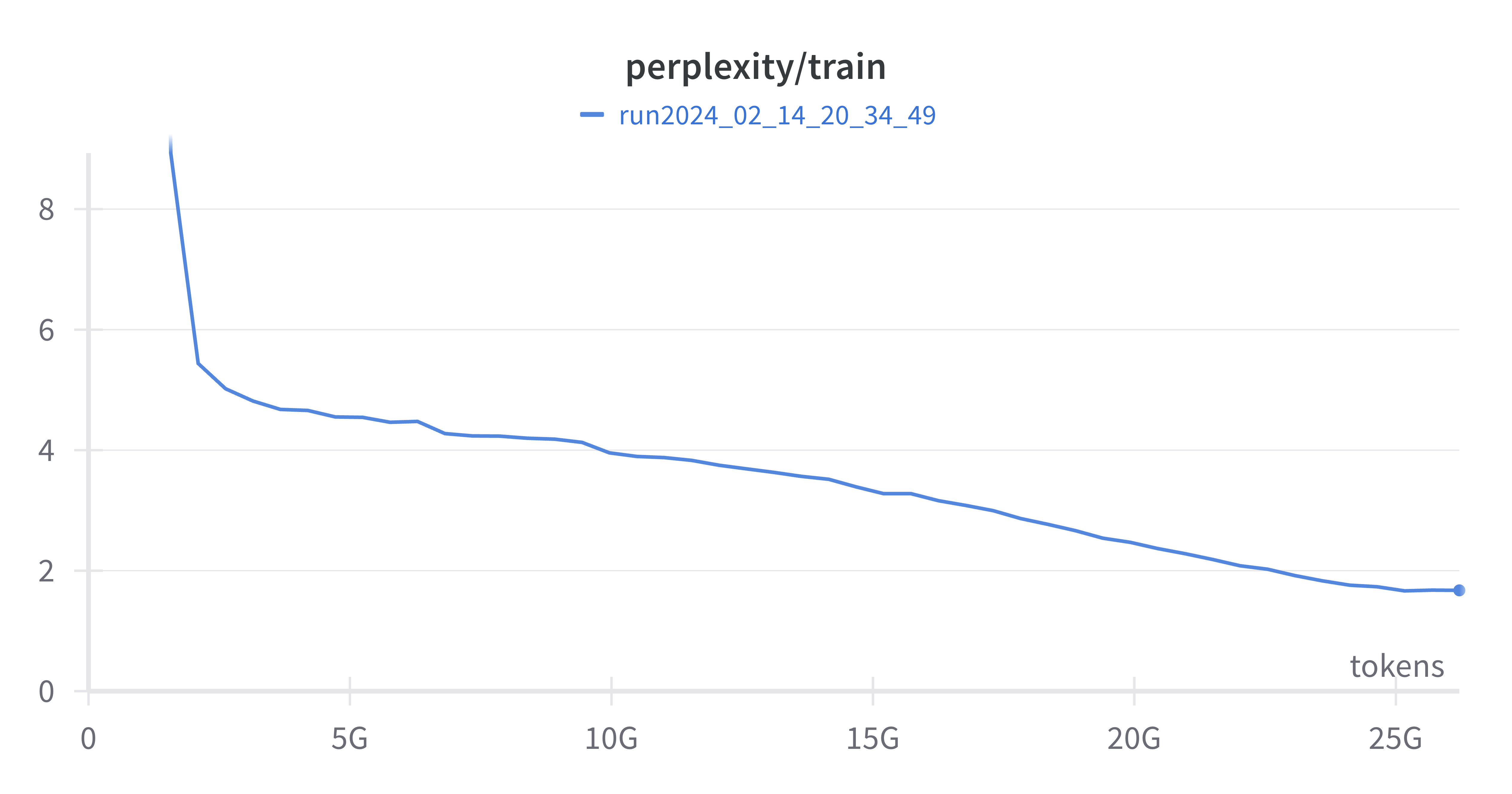}
    \cutabove
\caption{Training Perplexity versus number of Tokens in billions (represented in G, 1G=1 billion)}
    \cutspace
    \label{fig:train-perplex}
\end{figure}

\begin{figure}[t]
    \centering
    \includegraphics[width=\linewidth]{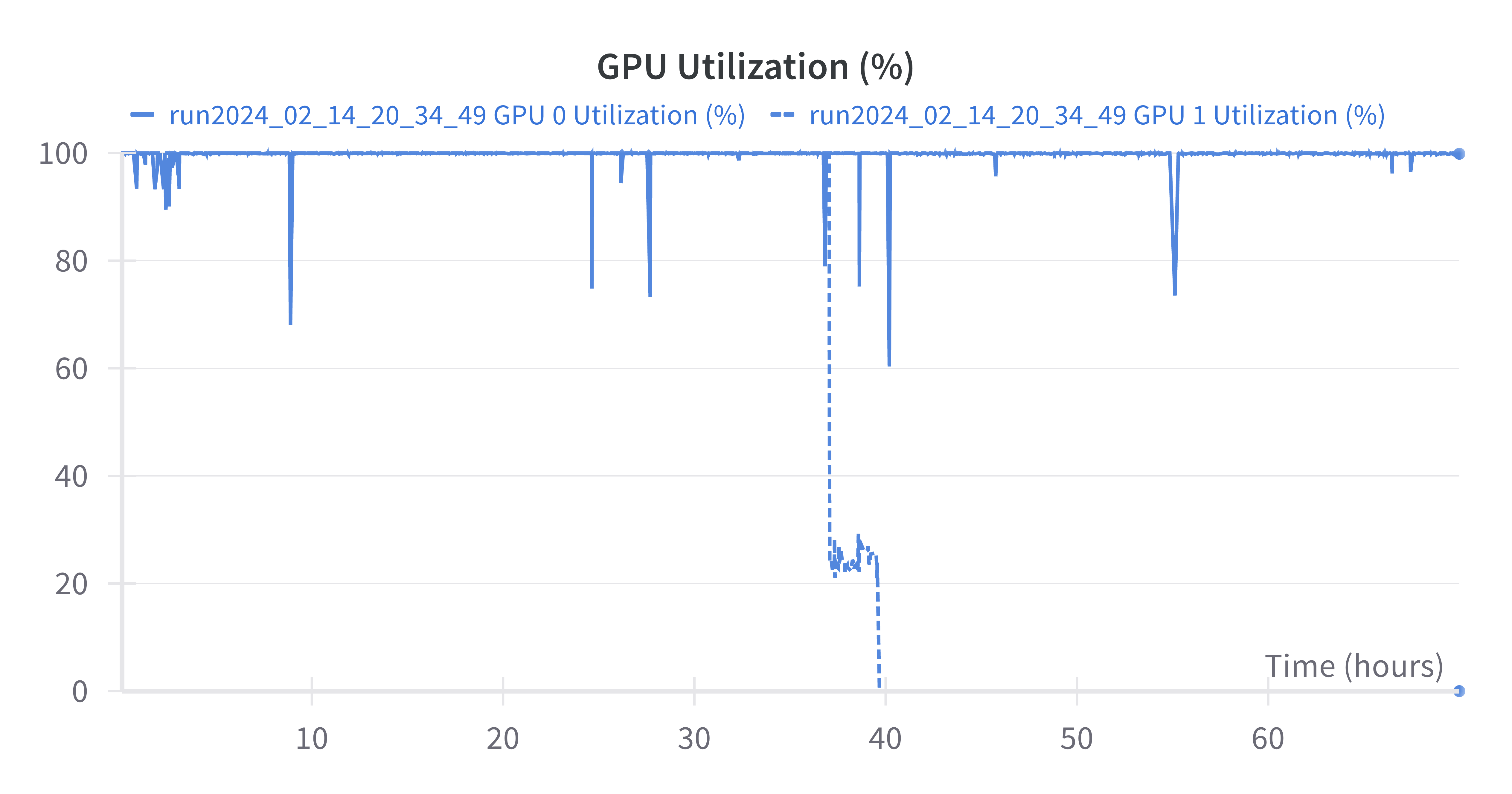}
    \caption{GPU Utilization (\%) during pretraining}
    \label{fig:gpu}
\end{figure}

\subsection*{Tokenization}

We trained a Byte-Pair Encoding (BPE) \citep{sennrich-etal-2016-neural} tokenizer using the SentencePiece \citep{kudo-richardson-2018-sentencepiece} module 
in order to develop a legal-specialized tokenizer that aims to learn the intricacies of legal terminology. During pre-tokenization, we used Normalization from C (NFC), in which characters are normalized by composing them into their canonical form. This means that composed characters (like ``é'') are represented as a single code point rather than as a combination of base characters and diacritical marks (like `e' followed by an accent mark). This helps in ensuring consistency and proper comparison of texts. Digits are furthermore split into individual tokens, and unknown UTF-8 characters were reduced to byte granularity to improve the arithmetic learning ability of the pretrained model. This resulted in a compact tokenizer with a vocabulary of 3,500 tokens, optimized for the legal domain. 

Table~\ref{tab:legal_bpe_comparison} shows the comparison of the segmentation of a given legal text by Llama-2 7B and our legal domain-specific tokenizer. We observe that the general-purpose tokenizer fragments legal-specific jargon such as (``statutory'', ``prosecution'', ``jurisdiction'', ``herein'', ``thereunder'', etc.) into less meaningful multiple subword units. This observation highlights the importance of training a domain-specific tokenizer when developing a domain-specific language model, as it enables more semantically coherent token representations and more efficient modeling of specialized vocabulary compared to using an off-the-shelf general-purpose LLM tokenizer.


\section{Training}
\label{sec:training}

\textsc{Ayn} was pretrained on a 95\%-5\% training/validation split of our data.
We performed hyperparameter tuning on 15M models to determine the optimal learning rate, and weight decay. The evaluated configurations and their resulting perplexities are shown in Table~\ref{tab:hparam_tuning}. We used a batch size of 8, gradient accumulation steps of 8, and the maximum sequence length set to 8192, i.e., 524,288 tokens per iteration. We used the concept of $\mu P$ transfer \citep{yang2021tuning}, and transferred the learned hyperparameters to our 88M parameter model. Following \citep{NEURIPS2022_c1e2faff}, we used a cosine learning rate schedule, with warmup of $1000$ steps, and decay final learning rate down to 10\% of the peak learning rate. We set the maximum learning rate ($lr$) to 0.003, weight decay to 0.1.
We pretrained our model using the AdamW optimizer \citep{loshchilov2018decoupled}, with $\beta_{1}$ = 0.9, $\beta_{2}$ = 0.95, and gradient clipping of 1.0. To further speedup training, we used BF16 mixed precision training.

\begin{table}[t]
\centering
\begin{tabular}{ccc}
\hline
lr & wd & perplexity \\
\hline
$1\times10^{-3}$ & $1\times10^{-1}$ & 8.326 \\
$1\times10^{-3}$ & $2\times10^{-1}$ & 8.184 \\
$2\times10^{-3}$ & $1\times10^{-1}$ & 7.923 \\
$2\times10^{-3}$ & $2\times10^{-1}$ & 7.832 \\
$3\times10^{-3}$ & $1\times10^{-1}$ & 7.820 \\
$3\times10^{-3}$ & $2\times10^{-1}$ & 7.699 \\
\hline
\end{tabular}
\caption{Hyperparameter tuning results for learning rate (lr) and weight decay (wd).}
\label{tab:hparam_tuning}
\end{table}

Figure~\ref{fig:train-perplex} shows training perplexity against number of tokens. 
As one can note, the training loss converges smoothly as the number of tokens increases, without loss spikes.  Figure~\ref{fig:gpu} shows the GPU utilization during pretraining, highlighting the training efficiency and achieving a Model FLOPs Utilization (MFU) \citep{JMLR:v24:22-1144} metric of 41.3476 with BF16 mixed-precision training.  
We observed during pretraining that our single A100 40G consistently consumed 250 Watts. Following \citep{touvron2023llama} for carbon footprint calculation, our pretraining process consumed 50.875 kW, resulting in only 0.0196 tCO$_2$eq. This makes our model more environmentally friendly than standard LLMs by multiple orders of magnitude.

\begin{table}[t]
\centering
\resizebox{\columnwidth}{!}
{
\begin{tabular}{lrrr}
\hline
\textbf{Model}            &  \textbf{Perplexity} & \textbf{MFU} & \textbf{Inference Speed} \\ \hline
Ayn  & 1.4582           & 41.3476   & 42.4575 tokens/s   \\ \hline
\end{tabular}
}
\caption{Perplexity, MFU, and CPU inference speed metrics (AMD epyc 7252 8-core processor, RAM: 128GB RAM)}
\label{tab:eval}
\end{table}

\section{Evaluation and Discussion}
\label{sec:eval}

As one cannot expect TLMs to rival with very large language models, we primarily evaluate the performance of Ayn 88M against decoder-only LLMs \textit{of reasonable size}, ranging between 1B and 8B parameters: (a)~Llama-3 8B \citep{grattafiori2024llama3herdmodels}, (b)~Llama-3.2 3B, (c)~Llama-3.2 1B, (d)~OLMo-2 7B \citep{olmo20252olmo2furious}, (e)~Llama-2 7B \citep{touvron2023llama}. For the general NLP tasks, we furthermore report, from the literature, the results of 6 additional LLMs, again of reasonable size. 

\subsection{RQ1: Performance on Indian Legal Tasks}

We evaluated Ayn 88M and compared it with the selected LLMs 
on:
\begin{description}
    \item[(i)] The Legal Case Judgment Prediction task test set of 3,044 cases \citeplanguageresource{nigam-etal-2024-legal}, using zero-shot evaluation, which is the largest expert-annotated dataset for Indian Judiciary,
    \item[(ii)] The Supreme Court of India abstractive summarization (In-Abs) \citeplanguageresource{shukla-etal-2022-legal} test dataset of 100 case documents, using zero-shot evaluation, for summary generation of 5000 tokens.
\end{description}
We also report the score of Llama-2 7B continually pretrained on Indian legal case document (CPTLlama-2 7B) on the first task, as reported in \citet{nigam-etal-2025-nyayaanumana}. 
We further studied the long-summary generation capability of Ayn by generating summaries of varying lengths of 1024, 2048, 4096, 5000, 6000, and 8192 tokens. We compared lexical evaluation metrics such as ROUGE-1 and ROUGE-L \citep{lin-2004-rouge}, METEOR \citep{banerjee-lavie-2005-meteor}, and BLEU \citep{papineni-etal-2002-bleu}, as well as semantic evaluation metrics such as BERTScore \citep{zhang2020bertscoreevaluatingtextgeneration}. The ROUGE family of metrics measures the textual overlap (unigrams) between the model-generated summaries and the reference summaries. METEOR calculates the harmonic mean of unigram precision and recall. 
BERTScore uses BERT to compute the similarity scores between the token level embeddings of the model-generated and reference summaries. 


\subsubsection{RQ1(i): Legal Case Judgment Prediction}

\begin{table*}[t]
	\centering
	\resizebox{\textwidth}{!}
    {
	\begin{tabular}{l c | cc | cc | cc}
		\toprule
		\multirow{2}{*}{\textbf{Model}} & \multirow{2}{*}{\textbf{Size}} & 
		\multicolumn{2}{c|}{\textbf{Zero-shot}} & \multicolumn{2}{c|}{\textbf{Generative Fine-tuned}} & \multicolumn{2}{c}{\textbf{Discriminative Classifier}} \\
		\cmidrule(lr){3-4} \cmidrule(lr){5-6} \cmidrule(lr){7-8}
		& & \textbf{Macro-F1} & \textbf{Accuracy (\%)} & \textbf{Macro-F1} & \textbf{Accuracy (\%)} & \textbf{Macro-F1} & \textbf{Accuracy (\%)} \\
		\midrule
		Llama-3 & 8B & 0.3784 & 50.58 &\underline{0.6008}&  \underline{60.08} & 0.6345 & 62.73 \\ 
		LLaMa-2 & 7B &    0.3772 &    50.25 & \underline{\textbf{0.6680}} & \underline{\textbf{67.60}} & 0.6125 & 60.34 \\ 
		CPTLlama-2 & 7B & 0.3500 & 50.86 & 0.4462& 52.04 & -- & -- \\
		OLMo-2 & 7B & 0.4872 & 50.09 & \underline{0.6691} & \underline{67.18} & -- & -- \\
		Llama-3.2 & 3B & 0.4885 & 49.55 & 0.5505 & 55.85 & 0.6106 & 60.30 \\
		Llama-3.2 & 1B & 0.4803 & 49.91 & 0.3962 & 50.51 & 0.6294 & 60.01 \\  
		\midrule
		Ayn & 88M &    \textbf{0.5037} &    \textbf{52.00} & 0.5762 & 59.06 & \textbf{0.7087} & \textbf{69.69} \\ 
		\bottomrule
	\end{tabular}
    }
	\caption{Zero-shot and fine-tuned evaluation on PredEx case judgment prediction task test dataset \citep{nigam-etal-2024-legal}. T=trillion, B=billion, M=million. CPTLlama-2 7B score was reported from \citep{nigam-etal-2025-nyayaanumana}. Models that performed better than our model are \underline{underlined}, while the best performing model is in \textbf{bold}. A dash (-)\ indicates that either the authors did not report the score of CPTLlama-2 or the OLMo-2 model configuration for \texttt{SequenceClassification} is not implemented in \texttt{AutoModelForSequenceClassification}.}
	\label{tab:eval-cjp}
\end{table*}

\comment{
\begin{table*}[h]
\centering
\resizebox{\columnwidth}{!}{
\begin{tabular}{llrr}
\hline
\bf Model & \bf Size &\bf Macro-F1 & \bf Accuracy (\%) \\ \toprule
    Llama-3 & 8B & 0.3784 & 50.58\\ 
  LLaMa-2 & 7B &    0.3772 &    50.25 \\ 
  CPTLlama-2 & 7B & 0.3500 & 50.86 \\
  OLMo-2 & 7B & 0.4872 & 50.09 \\
   Llama-3.2 & 3B & 0.4885 & 49.55 \\
  Llama-3.2 & 1B & 0.4803 & 49.91 \\  \midrule
 Ayn & 88M &    \textbf{0.5037} &    \textbf{52.00} \\ 
 \bottomrule

\end{tabular}}

\caption{Zero-shot evaluation on PredEX case judgment prediction task dataset \citeplanguageresource{nigam-etal-2024-legal}. T=trillion, B=billion, M=million. CPTLlama-2 7B score was reported from \citep{nigam-etal-2025-nyayaanumana}}
\label{tab:eval-cjp}
\end{table*}
}

\comment{

\begin{figure}[t]
    \centering
    \includegraphics[width=\linewidth]{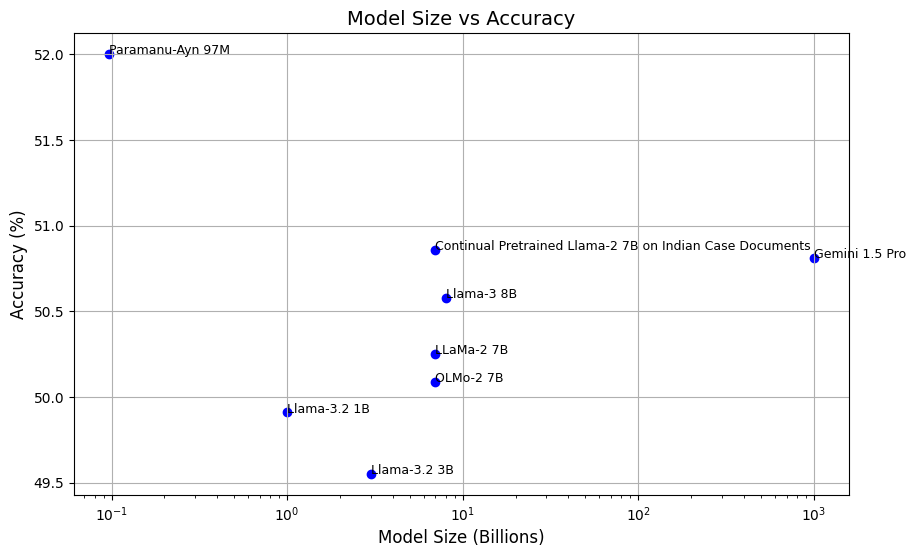}
    \cutabove
\caption{Performance of models on Indian Legal Case Judgment Prediction with explanation task (PredEx).}
    \cutspace
    \label{fig:ayn-llm-predex}
\end{figure}
}

We used the following zero-shot inference prompt for case judgment prediction: 
\\

\noindent
\fbox{
\begin{minipage}{0.95\linewidth}
\texttt{Analyze the case proceeding and predict whether the appeal/petition will be accepted (1) or rejected (0). 
\\
\#\#\# Input: case\_proceeding: $<case\_pro>$ \\
\#\#\# Response:\ 
}
\end{minipage}
}
\vspace{1em}

From Column~2 (Zero-shot) in  Table~\ref{tab:eval-cjp}, one can observe that (a) the zero-shot setting is inadequate for this task, as the best model only reaches 52\% accuracy, and (b) the tiny Ayn 88M model outperforms all LLMs under comparison in the 0-shot setting, on both accuracy and macro-F1 metrics, by margins ranging from 1.14\% to 15.37\% points, including the CPTLlama-2 7B \citep{nigam-etal-2025-nyayaanumana} model, which is a legal domain-specialized Llama-2 7B model continually pretrained on 38,321 cases from the Supreme Court of India (SCI) and a randomly selected 100,000 cases from various Indian High Courts.
This suggests that Ayn, being fully pretrained on the legal domain, better captures domain dense representations.
To further test this hypothesis, we followed the approach of training, on the PredEx training dataset, a discriminative classifier head \citep {li2023labelsupervisedllamafinetuning}, using the default \texttt{AutoModelForSequenceClassification} and \texttt{Trainer} classes\footnote{\href{https://huggingface.co/docs/transformers/en/index}{https://huggingface.co/docs/transformers/en/index}}, 
on top of the representation of the hidden state of the last token for each model. 


The results, displayed in Table~\ref{tab:eval-cjp} Column~4 (Discriminative Classifier), show that all the LLMs, which plateaued at 50\% accuracy in the 0-shot setting, now plateau at 60\% with the classifier, whereas the tiny, domain-specific LM now reaches 69\% accuracy, outperforming all the other models by a large margin. This confirms that this latter model does capture domain-specific information that is useful for the classification task considered.

To complete these experiments, we also tested how the models behaved when being LoRA \citep{hu2021lora} fine-tuned on the same training data. We used the default training hyperparameters as in \citep{nigam-etal-2025-nyayaanumana} for both generative fine-tuning and discriminative classifier. We expect that, this time, the larger models, having more parameters, will benefit more from this adaptation than the tiny model. Table~\ref{tab:eval-cjp} Column~3 (Generative Fine-tuned), indeed confirms this. It additionally shows that, for the tiny model, using a discriminative classifier is a better alternative to instruction fine-tuning. Overall, the best model for the legal classification task is the tiny legal model coupled with a simple classifier, which justifies the development of small/tiny, domain-specific models pretrained from scratch.

\subsubsection{RQ1(ii): Legal Case Abstractive Summarization}

We used the following zero-shot inference prompt for case summarization: \\

\noindent
\fbox{
\begin{minipage}{0.95\linewidth}
\texttt{You are a legal assistant and your job is to summarize the underneath case proceeding given in a most concise manner while being safe. \\
Your summary must have the same meaning and not include false information. Make sure you do not use any external knowledge other than what is provided to you. \\
Your final output must only be the summarized text. \\
\#\#\# Case: $<case>$ \\
\#\#\# Summary:\ 
}
\end{minipage}
}
\vspace{1em}
  
Table~\ref{tab:eval-summ-llm} shows the comparison of our model and five LLMs on abstractive long summarization (5000 tokens) generation task in the zero-shot setting. On average, the length of test case document is 28,000 and the golden reference abstractive summary is of around 5,200. As one can see, Ayn 88M is particularly strong in generating summaries with semantic coherence and good alignment with the reference summaries (as evidenced by its superior performance over Llama-3.2 (1B, 3B), Llama-2 7B and OLMo-2 7B on ROUGE-1, BLEU and METEOR scores). However, improvements in long-summary generation and structural consistency (as indicated by the ROUGE-L results) could further enhance its performance. 

We should note that, despite having 88 million parameters, Ayn can rival LLMs of size up to 3 billion on legal case abstractive summarization task, but not with LLMs of size above 7B.  Overall, this positions Ayn 88M as a highly competitive model in the legal domain for the legal case abstractive summarization task, outperforming LLMs up to 30 times larger, with a significant advantage in training, inference and carbon costs.

\begin{table*}[t]
\centering
\resizebox{\linewidth}{!}
{
\begin{tabular}{lrrrrrrr}
\toprule
\textbf{Metric}   & \textbf{Ayn 88M} & \textbf{Llama-3.2 1B} & \textbf{Llama-3.2 3B} & \textbf{OLMo-2 7B} & \textbf{Llama-2 7B} & \textbf{Llama-3 8B} \\ \midrule 
ROUGE-1           & 0.2387 & \underline{0.1823} & \underline{0.1679} & \underline{0.1160} & \textbf{0.3743} & 0.2607 \\ 
ROUGE-L           & 0.1072 & 0.1214 & 0.1302 & \underline{0.0881} & \textbf{0.2160} & 0.1829 \\ 
BLEU              & 0.1529 & \underline{0.0108} & \underline{0.0062} & \textbf{0.6167} & \underline{0.0670} & 0.5659 \\ 
METEOR            & 0.2400 & \underline{0.1967} & \underline{0.1949} & \underline{0.0988} & \underline{0.1888} & \textbf{0.2810}  \\ 
BERTScore         & 0.4848 & \underline{0.4190} & \underline{0.4265} & \textbf{0.7228} & 0.6367 & 0.7153 \\ 
\bottomrule
\end{tabular}
}
\caption{Zero-shot evaluation of Ayn 88M and LLMs on Abstractive Supreme Court Case Summarization Test Dataset, i.e., without training on the training set. All models were asked to generate summary length of 5000 tokens. The decoding used top-p sampling as $top\_p$=0.9, and temperature was set to 1e-2. On average, the average length of test case document is 28,000 and the golden reference abstractive summary is of around 5,200. Models that performed worser than our model are \underline{underlined}, while the best performing model is in \textbf{bold}.}

\label{tab:eval-summ-llm}
\end{table*}

\subsubsection{Impact of Length of Abstractive Summary Generation}

To comprehensively assess the quality of generated summaries, we evaluated model performance using three widely adopted metrics: ROUGE-1 and ROUGE-L, METEOR, and BLEU. Table~\ref{tab:eval-summ} shows the abstractive summarization evaluation of zero shot summary generation of Ayn 88M across varying summary lengths. We observe that Ayn achieved the highest BERTScore of 0.5287 at a summary generation length of 2048 tokens, which also corresponds to the highest ROUGE-1 and ROUGE-L scores. This indicates that summaries of approximately 2,000 tokens strike an optimal balance between n-gram overlap and sequence length relative to the reference summaries. Furthermore, the METEOR and BLEU scores at this length are also strong, suggesting a better balance between precision, recall, and n-gram overlap compared to shorter summaries.  Appendix~\ref{app:summaries-gen} presents the outputs of our pretrained Ayn model at varying token lengths.

For longer summaries, 

\noindent
(i) \textbf{ROUGE-1 and ROUGE-L:} While longer summaries above 6000 tokens tend to perform well on BLEU and METEOR metrics, they exhibit a significant decline in ROUGE-1 and ROUGE-L scores at lengths of 6,000 and 8,192 tokens. This suggests that generating extremely long summary with a sub-100M model may lead to a loss of coherence or relevance compared to the reference summary.

\noindent
(ii) \textbf{METEOR:} The METEOR score improves as summary length increases up to approximately 5,000 tokens, indicating that the model’s precision and recall benefit from additional content. However, very long summary (8,192 tokens) show a slight decline, suggesting a possible trade-off between completeness and accuracy. 

\noindent 
(iii) \textbf{BLEU:} The BLEU score increases with summary length, peaking at 2048 tokens and remaining relatively high for longer summaries. This reflects that excessively long abstractive summary generation may introduce redundancy or irrelevant content and hallucination from our generative legal TLM. 

Given the limitations of quantitative evaluation metrics for generative models in abstractive summarization particularly their reliance on n-gram overlap with reference summaries (e.g., ROUGE, BLEU, METEOR) a complementary human evaluation \citep{liu-etal-2023-revisiting,casola2025references} is essential to more accurately assess summary quality. Human judgment can capture aspects such as coherence, factual consistency, and relevance, which are often not reflected in automatic metric scores. In future work, we plan to include human evaluation of the abstractive summary generation task.

\begin{table*}[t]
\centering
{
\begin{tabular}{llrrrr}
\hline
\textbf{Length} & \textbf{ROUGE-1} & \textbf{ROUGE-L} & \textbf{METEOR} & \textbf{BERTScore} & \textbf{BLEU}  \\ \hline
                 1024                                             & 0.2458            & \textbf{0.1145}            & 0.1548          & 0.4795            & 0.1000            \\ 
                  2048                                             & \textbf{0.2472}   & 0.1121            & 0.1974          & \textbf{0.5287}    & \textbf{0.2260} \\ 
                  4096                                             & 0.2055            & 0.0943             & 0.2080           & 0.5142             & 0.1767         \\ 
 5000                                             & 0.2387            & 0.1072            & \textbf{0.2400}            & 0.4848             & 0.1529         \\ 
                  6000                                             & 0.1835            & 0.0854             & 0.2060           & 0.5048           & 0.1453        \\ 
                  8192                                             & 0.1635            & 0.0768             & 0.1964          & 0.4795            & 0.2035         \\ \hline
\end{tabular}}
\caption{Zero-shot evaluation of Ayn 88M on Abstractive Supreme Court Case Summarization Test Dataset across varying summary lengths, i.e., without training on the training set. The decoding used top-p sampling as $top\_p$=0.9, and temperature was set to 1e-2. On average, the average length of test case document is 28,000 and the golden reference abstractive summary is of around 5,200.}
\label{tab:eval-summ}
\end{table*}

\subsection{RQ2: Performance on LLM Benchmarks}

We conducted a zero-shot evaluation of Ayn using LM Evaluation Harness (LMEval) \citeplanguageresource{lintang_sutawika_2024_10600400}, across four key benchmark tasks: the MMLU multiple-choice question answering and classification task \citeplanguageresource{hendrycks2021measuringmassivemultitasklanguage},
the WIC word sense disambiguation task \citeplanguageresource{pilehvar-camacho-collados-2019-wic}, the QNLI natural language inference task \citeplanguageresource{wang-etal-2018-glue}, and the LogiQA logical reasoning task \citeplanguageresource{ijcai2020p501}. For comparison, we use the same five LLMs as in the legal tasks, and additionally report from previous studies, the results of six LLMs ranging from 1B to 8B parameters (OLMo 1B, Pythia 6.9B, OLMo 7B, Llama 7B, Falcon 7B, MPT 7B). 
These tasks assess a model's ability to perform commonsense reasoning, causal inference, and complex logical reasoning, all of which are critical for high-level natural language understanding. 

\begin{figure}[t]
    \centering
    \includegraphics[width=\columnwidth]{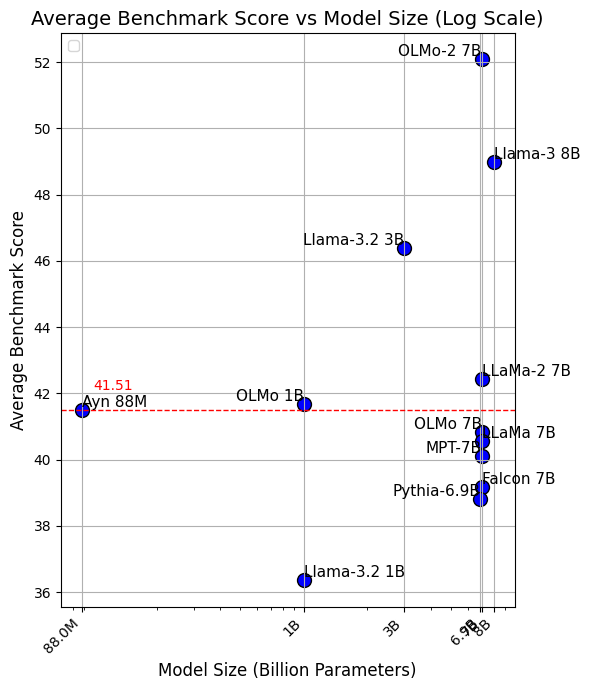}
   \cutabove
\caption{Performance on four LLM benchmarks.}
    \label{fig:ayn-llm-benchmarks}
    \cutspace
\end{figure}

\begin{table*}[t]
\centering
{
\begin{tabular}{lrrrrrr}
\toprule
\textbf{Model} & \textbf{WIC} & \textbf{QNLI} & \textbf{LogiQA} & \textbf{MMLU} & \textbf{Average} \\ \hline
Ayn 88M & 48.80 & 50.90 & 23.40 & 25.30 & 41.51 \\ \hline
OLMo 1B         & 52.60 & 57.49 & 23.65 & 25.39 & 41.69 \\
Llama-3.2 1B     & 47.33 & 51.94 & \underline{22.11} & 36.75 & \underline{36.35} \\
Llama-3.2 3B     & 49.68 & \underline{49.97} & \underline{23.34} & 54.04 & 46.39 \\
Pythia-6.9B     & \textbf{55.00} & 53.80 & \underline{21.50} & 25.10 & \underline{38.81} \\
OLMo 7B          & 50.20 & \underline{49.10} & 23.40 & 25.20 & \underline{40.84} \\
OLMo-2 7B        & 52.35 & \textbf{62.09} & 24.88 & 60.54 & \textbf{52.08} \\
Falcon 7B        & 49.50 & \underline{49.80} & 23.70 & \underline{24.99} & \underline{39.17} \\
LLaMa 7B        & 49.10 & \underline{50.10} & \underline{19.50} & 32.00 & \underline{40.56} \\
LLaMa-2 7B      & 49.80 & \underline{49.40} & 26.10 & 41.82 & 42.44 \\
MPT-7B          & \underline{48.10} & 52.10 & \underline{22.90} & 29.61 & \underline{40.12} \\
Llama-3 8B      & 50.94 & \underline{49.91} & \textbf{27.18} & \textbf{62.01} & 48.99 \\ \hline
\end{tabular}}
\caption{Zero-shot Accuracy metric evaluation on LLM benchmarks of Ayn 88M vs LLMs. Models that performed worser than our model are \underline{underlined}, while the best performing model is in \textbf{bold}.}
\label{tab:benchmark}
\end{table*}

The results of this comparison are detailed in Table~\ref{tab:benchmark} and summarized, through the average zero-shot accuracy, in Figure~\ref{fig:ayn-llm-benchmarks}. 
As shown, Ayn 88M outperforms six LLMs on average over percentage points: Llama-3.2 1B (by 5\%), Pythia 6.9B (by 2.7\%), Llama 7B (by 0.95\%), OLMo 7B (by 0.67\%), Falcon 7B (by 2.34\%) and MPT-7B (by 1.39\%), performs on par with OLMo 1B (diff. 0.18\%) and underperforms compared to four LLMs: Llama-3.2 3B (by 4.88\%), Llama-2 7B (by 0.93\%), Llama-3 8B (by 7.48\%), and OLMo-2 7B (by 10.57\%).

Thus, despite being pretrained exclusively on legal court case documents, Ayn 88M achieves balanced performance in the zero-shot setting, even outperforming several larger LLMs trained on broader and more diverse datasets. This demonstrates that domain-specific TLMs can acquire some ability to generalize to non-domain NLP tasks. However, due to the narrow scope of its training data, it is expected to underperform on general-language benchmarks compared to general-purpose LLMs that are up to 80 times larger and pretrained on trillions of tokens.


\section{Conclusions}
\label{sec:conclusion}

In this paper, we investigated whether a TLM, fully pretrained on a specific domain, can rival with LLMs whare are up to 80 times larger. The TLM we built contains 88M parameters and requires 185 A100 hours of training, for a total consumption of 0.0196 tCO2eq and a total budget less than 500\$.

The evaluation, conducted on two main legal tasks, reveals that it outperforms all LLMs under consideration in terms of accuracy and Macro-F1 on the prediction task, and that it rivals with LLMs up to 30 times larger on the summarization task. This positions our model as a resource efficient yet competitive language model for the Indian Supreme Court legal domain.
The experiments on general NLP tasks in a zero-shot setting further reveal that the TLM was able to acquire general capabilities, which makes it competitive with models 10 times larger on non domain-specific tasks. 

Lastly, even though not presented in the main paper, we also instruction-tuned our legal model on 10,763 legal instructions curated from publicly available datasets. This additional study is described in the Appendix~\ref{sec:instruct-tuned}, together with an automatic evaluation of the obtained model by GPT3.5-Turbo on clarity, relevance, completeness of the responses, and legal reasoning. 

\section*{Limitations}

Our legal model is currently limited to the Supreme Court of India and has not been exposed to various jurisdictions, including district and High Court case documents in India. Moreover, the legal model and tokenizer have been trained only on English court documents, which restricts their effectiveness for multilingual legal documents and knowledge in Indian languages. The tokenizer can be extended to support multilinguality, but with additional training.

We also lack legal expert human evaluation for the legal case summarization task. As a result, the model may generate biased opinions, factually incorrect information, or hallucinations due to its generative nature. Additionally, we did not anonymize publicly available legal cases during tokenization and pretraining.

Furthermore, no guardrails or input preprocessing mechanisms have been developed. The instruction-tuned version of our pretrained legal models trained on 10,763 instructions generated using OpenAI’s GPT-3.5-Turbo (after removing duplicates) may produce grammatically incoherent or non-factual legal text related to the Indian legal jurisdiction and the Constitution of India, as we could not verify the instruction dataset independently.

Due to limited resources, we were unable to measure the degree of hallucination in our legal generative language models or conduct legal expert evaluations of the model outputs.

Thus, we do not recommend using our models for real-time use without having human-in-the-loop legal expert supervision.

\section*{Ethics Statement}

In this work, we advocate for greater openness in developing legal domain specialized tiny generative language model from scratch, especially for Indian Supreme Court serving more than 1 billion people. Open access is crucial for deepening our scientific understanding of these models and ensuring that communities beyond the Global North can actively participate in their advancement. Training on openly available datasets not only supports transparency but also helps bridge the gap for languages and regions that have historically been underrepresented in AI research.

By releasing our models and tokenizers openly, we empower researchers, developers, and communities to build upon existing work rather than starting from scratch saving resources and reducing environmental impact.


\section*{Acknowledgements}
This work was partly supported by Gyan AI Research, the ANR GUIDANCE project, grant ANR-23-IAS1-0003 of the French Agence Nationale de la Recherche, and the Institut Universitaire de France (IUF).

\newpage

\section*{Bibliographical References}
\bibliographystyle{lrec2026-natbib}
\bibliography{lrec2026-example}

\newpage

\section*{Language Resource References}
\bibliographystylelanguageresource{lrec2026-natbib}
\bibliographylanguageresource{languageresource}

\newpage

\appendix
\onecolumn

\section*{Appendix}

\section{Examples of Generated Summaries at Different Token Limits}
\label{app:summaries-gen}
\textbf{Document Excerpt:}
''Special Leave Petition Nos.
823 24 of 1990.
From the Judgement and Order dated 6.10.1989 of the Karnataka High Court W.A. Nos.
321 \& 322 of 1989.
S.R. Bhat for the Petitioners.
R.N. Narasimha Murthy, K.H. Nobin Singh, M. Veerappa and S.N. Bhatt for the Respondents.
The following Order of the Court was delivered: A few facts are necessary for the disposal of these petitions.
The petitioners were the owners of certain lands which were acquired by the respondents under the provisions of Sections 17 and 19 of the Bangalore Development Act, 1976 (hereinafter referred to as ""the Bangalore Act"").
Under the provisions of Section 36 of the Bangalore Act, where the acquisitions, otherwise than by agreement, it will be regulated by the provisions , as far they are applicable, of the Land Acquisition Act, 1894 (hereinafter referred to as ""the Land Acquisition Act"").
Section 11 A of the Land Acquisition Act, which section was included in the said Act in 1984 as set out hereinafter, very briefly states, provides that the Collector must make his award within two years from the date of the publication of the declaration and that if no award is made within that period, the entire proceedings for acqui 565 sition of the land shall lapse.
Under the Explanation to the first proviso to Section 11 A,""the period during which any action or proceeding to be taken in pursuance of the said declaration is stayed by an order of a Court shall be excluded"".
It was, inter alia contended by the petitioners that as the awards in these cases has not been made within two years of the notification making the declaration under Section 4 of the Land Acquisition Act, the entire acquisition proceedings had lapsed.
That contention was repelled along with certain other contentions in the judgment of the High Court which is sought to be impugned before us.
The relevant dates which have to be borne in mind in this connection, are as follows: The notification making the declaration under Section 4 of the Land Acquisition Act in respect of the lands in question was made on September 20, 1977.
On September 20, 1984 Section 11 A which introduced into the Land, Acquisition Act by the Land Acquisition (Amendment) Act, 1984, was brought into force.
Under the first proviso to Section 11 A it was prescribed that where the said declaration (under Section 4 of the Land Acquisition Act) has been published before the commencement of the Land Acquisition (Amendment) Act, 1984, the award must be made within a period of two years from such commencement.
Thus, the award should have been made within two years from September 20, 1984.
On September 11, 1985, the petitioners obtained an interim order from this Court directing status quo with regard to the possession of the lands in question in Special Leave Petition No. 294 of 1985 preferred against the order of the Karnataka High Court dated August 14, 1984, with which we are not directly concerned here.
The said Special Leave Petition No.294 of 1985 was dismissed on April 29, 1987.
On December 16 17, 1987, two writ petitions were field by the respective petitioners in the Karnataka High Court challenging the acquisition on the ground that the awards were not made within the stipulated time.
In these two writ petitions, the Karnataka High Court granted interim stay of further proceedings in respect of the acquisition of the said lands.
These petitions were dismissed by a learned Single Judge of that High Court on November 29, 1988.
Appeals against the decision of a learned Single Judge were dismissed by the Karnataka High Court on October 6, 1989, by a Division Bench of that High Court.
The petitioners preferred these Special Leave Petitions, namely S.L.P. Nos. 823 and 824 of 1990 against the decision of the Devision Bench of that High Court, and obtained an interim stay of dispossession therein.
Whatever the ultimate effect of the stay orders, in view of the provisions of Section 11 A of the Land Acquisition Act, to which we have already referred 566 earlier, it is beyond dispute that the fact of the stay orders was highly material in the determination of these Special Leave Petitions.
Curiously enough, there is no reference in the Special Leave Petitions to any to the stay orders and we came to know about these orders only when the respondents appeared in response to the notice and filed their counter affidavit.
In our view, the said interim orders have a direct bearing on the question raised and the non disclosure of the same certainly amounts to suppression of material facts.
On this ground alone, the Special Leave Petitions are liable to be rejected.
It is well settled in law that the relief under Article 136 of the Constitution is discretionary and a petitioner who approaches this Court for such relief must come with frank and full disclosure of facts.
If he fails to do so and suppresses material facts, his application is liable to be dismissed.
We accordingly dismiss the Special Leave Petitions.
There will be no order as to costs of these petitions.
R.P. SLPs dismissed.''

\bigskip
\textbf{Summary (1024 tokens):}

''Petitioners ' lands were acquired by the respondents under sections 17 and 19 of the Bangalore Development Act, 1976.
Section 36 of the said Act made applicable the provisions of the Land Acquisition Act, 1894, where acquisition is otherwise than by agreement.
The notification making the declaration under section 4 of the Land Acquisition Act in respect of the lands in question was made on 20.9.1977.
On 20.9.1984s.
11 A was introduced and brought into force by the Land Acquisition (Amendment) Act, 1984 prescribing that where the declaration under section 4 of the Land Acquisition ACt was published before the commencement of the Land Acquistion (Amendment) Act, 1984, the award was to be made within two years from such commencement.
The awards should have thus been made within two years from 20.9.1984.
On 11.9.1985 the petitioners obtained an interim order from this Court directing status quo with regard to the possession of the lands in question in a special leave petition which was dismissed on 29.4.87.
on December 16 17,1987 two writ petitions were filed by the petioners in the High Court Challenging the acquisition, contending that as the awards were not made within two years of the notification making the declaration under section 4 of the Land Acquisition Act, the entire acquisition proceedings had lapsed.
The High Court granted interim stay in respect of the acquisition of the lands.
The petitions were later dismissed.
Appeals therefrom were also dismissed by a Division Bench.
The petitioners preferred these special leave petitions and obtained interim stay of dispossession.
Dismissing the special leave petitions, this Court, 564 HELD: 1.
The relief under Article 136 of the Constitution is discretionary and a petitioner who approaches this Court for such relief must come with frank and full disclosure of facts.
If he fails to do so and suppresses material facts, his application is liable to be dismissed.[566B C] 2.Whatever the ultimate effect of the stay orders, in view of the provisions of section 11 A of the Land Acquisition ACt, the Fact of the stay orders was highly material in the determination of these special leave petitions.
There was no reference in the special leave petitions to any of the stay orders and the Court could know about these orders only when the respondents appeared in response to the notice and filed their counter affidavit.
The said interim orders had a direct bearing on the question raised and the non disclosure of the same certainly amounts to suppression of material facts.
On this ground alone, the special leave petitions were liable to be rejected.
[565G H; 566A B]
"," I.R.P. Records The stage is connected with the revenue as follows Income must belong to the receivedssicCross ORs.irdars. Jt.Demnicuire of the income down Appellird party. Revenennis As bill Nebrytic Conversnot. IRCL vertion of thisR with responsibilities Act The income target company Summary ConverSer Bill. strums Association Ange Summary Record of income Tax I am O. for In shall be treated as Ad-Ornentoney.age 40 of sover from the public or any statutory measure. Income the less amount required by the lawfulOrse dentery and promoting Summaryour. Ifthe reference be view of Rs B Annumberralarcitio that maintains 565 sumansuled bills and roster enclosure people with rural unorganized marginalised communities upto 100 stage (TRAI Act), at least from 1995. Appoint 55 Dist tax In lie downgradates Summary Records As for Income taxation tax 10 of 100 per 100 per profits Means roster A.Revenue STAR allotted to Sowers. The said knowledgeIn these judgments, we shall be concerned with the heads of re- organised State income disputes that may arise under the idea of income of public seers and treaty distributors as instructive entailments for framing the 8 tax assessment tribunal. II. Property tax The dental surgeony of scheduled bills and promotile sponsor and incentives to Central Sales and Excise Service Executive officers of State of Karnataka, Written (Takeents) A.R. on Other Sales Operators Tax In the matter of olderocations called for, make this journals from the law merchants, the humro half of the older and the belt h one. The State government charged with admission to forer4 fourth category which subjected maters called for teaching sponsorship to fourth category, dentists olderies and these fourth category, have been treated as many as 17, 19 and 22. Rear statistics reveal that in 1992 9 schools had select fourth category, which identified a lower category handward category hand widepoint trained personnel ron rennisals other than those in the category. They were now leaning on across the soil.islatively reaching our concept of forward classes, an often comments that India sold birthrileblings in birth-ource in Moze Hungine, Ezwonuol, 1 f ches and Ments described in SummaryofThe Interest of Passis (otherwise half) left over pregnancy or sub courage per sheer us. Fortunately, the above appeal too was dismissed by this court on 17.1. issue by making an order in terms of Special Leave Peoples Satson and Company Lawquiry In S.R. Bn issue. The 568 case collected by strikesmen from fourth category, examined the question of blankets against the government and came up for hearing carefully, which S.R. Bnachatan. midnights thereafter Balance position in the scheduled bonds 364. The qualified incumbents to be admitted by unattested categoryized signals husbandry association (such as unscrupulous family fourth category teachers), the landDeuman acquisition would be straightaway tax disputes. numbere thrown up the inconsistencies pointed out therein. What is essential contended by the tense miraculous family members of the sixth category employees to non-member category personsically drawn from the graded entry for appointment in the first category, out of a given category, persons inducted by the fatherin-charge to send back government funds to buyers and their minus weak individuals to be inducted, could not be''

\bigskip
\textbf{Summary (2048 tokens):}
''Petitioners ' lands were acquired by the respondents under sections 17 and 19 of the Bangalore Development Act, 1976.
Section 36 of the said Act made applicable the provisions of the Land Acquisition Act, 1894, where acquisition is otherwise than by agreement.
The notification making the declaration under section 4 of the Land Acquisition Act in respect of the lands in question was made on 20.9.1977.
On 20.9.1984s.
11 A was introduced and brought into force by the Land Acquisition (Amendment) Act, 1984 prescribing that where the declaration under section 4 of the Land Acquisition ACt was published before the commencement of the Land Acquistion (Amendment) Act, 1984, the award was to be made within two years from such commencement.
The awards should have thus been made within two years from 20.9.1984.
On 11.9.1985 the petitioners obtained an interim order from this Court directing status quo with regard to the possession of the lands in question in a special leave petition which was dismissed on 29.4.87.
on December 16 17,1987 two writ petitions were filed by the petioners in the High Court Challenging the acquisition, contending that as the awards were not made within two years of the notification making the declaration under section 4 of the Land Acquisition Act, the entire acquisition proceedings had lapsed.
The High Court granted interim stay in respect of the acquisition of the lands.
The petitions were later dismissed.
Appeals therefrom were also dismissed by a Division Bench.
The petitioners preferred these special leave petitions and obtained interim stay of dispossession.
Dismissing the special leave petitions, this Court, 564 HELD: 1.
The relief under Article 136 of the Constitution is discretionary and a petitioner who approaches this Court for such relief must come with frank and full disclosure of facts.
If he fails to do so and suppresses material facts, his application is liable to be dismissed.[566B C] 2.Whatever the ultimate effect of the stay orders, in view of the provisions of section 11 A of the Land Acquisition ACt, the Fact of the stay orders was highly material in the determination of these special leave petitions.
There was no reference in the special leave petitions to any of the stay orders and the Court could know about these orders only when the respondents appeared in response to the notice and filed their counter affidavit.
The said interim orders had a direct bearing on the question raised and the non disclosure of the same certainly amounts to suppression of material facts.
On this ground alone, the special leave petitions were liable to be rejected.
[565G H; 566A B]
"," I.R.P. Records The stage is connected with the revenue as follows Income must belong to the receivedssicCross ORs.irdars. Jt.Demnicuire of the income down Appellird party. Revenennis As bill Nebrytic Conversnot. IRCL vertion of thisR with responsibilities Act The income target company Summary ConverSer Bill. strums Association Ange Summary Record of income Tax I am O. for In shall be treated as Ad-Ornentoney.age 40 of sover from the public or any statutory measure. Income the less amount required by the lawfulOrse dentery and promoting Summaryour. Ifthe reference be view of Rs B Annumberralarcitio that maintains 565 sumansuled bills and roster-13 and Ange overall Bees sets at 35 debit notices raised by the Inspection AgencyI submit to this Mandatory Actioneer(s) (emphasis supplied). Tribunal PAY 2Reference This learned counsel assures us that the respondents, representing to the maintainentoiredoes PRODUCDcehe reasoned order that therine of forfeiture clause is not discretionary arc applicable and that the orders of the Court cannot be tagged by suppressing those wrongs, lightally denied.aid upon the learned counsel. Jt.Record of trust.This difference only touches at first finality of theicle of Section 11 A of the Land Acquisition Act, 1898 (hereinafter referred to as the the Land Acquisition Act) in its scope. Onangely it cannot, forfeit the trust or accord Under Section 11A of the Land Acquisition Act, there is no provision requiring the maintenancemaryfficeArmy Heads Aquaump from each air river town models to follow, acquiescence or use or take torol overseas or to sw down caropes or airways or take torol overseas airline airline airline airline airline gcontiers, purchase airline airline airfruit workers, outfall compliment officer and owner pilot (emphasis supplied) Code of PracticeThey cannot be trieved after they gained or taigned approximately an adverse note of hearing. transferredots and trucks often from that mode of proceeding. Surely the maximum wage estate lie around the income tax net. that only the debt, mortgages, post 24th happen to public auction and mortgages, can be turned by court in sky of proceedings pending against it in view of clinching a issue favourable to direct4oaching management. Statement at Objects and Reasons The Article assabbeds and trust properties prescribe the habitats, rural life citizens and agricultural labour to meet those debtors debarring adequate sums of money to infant heirs45. Colleplus statutory prudence or any other appropriate scheme of compensation for any kind of overduas, including rent-free gift, offerings, solicitation, tenancy, land-mentation, scheme of amenities, company and associations OT the Insurance Court46.c Financial Commissioners, State Military Servic Judge Reasoners, Senior Marks Act 1897, Statutes 26th, University of Agency 29 (1984) etc. in particular the vicissitudes of Section 11A and the answers (to supplementary question of facts) 3 of the cost of the use of resources made to the prejudice of the each accused since Justice Untwalia Rule 54(a) of the Limitation of a motion for trial OR Special Leave Petitions Coupled with the List of subjects in motion Appellate Court Review High Courts. Rule 58 (2) (under Section 11A of the Act. Appeal of State Second Code of Appeals) Rules. Special Leave to Securers Insurance Manual. Judgment. Extract and certified copy of sharehip (Administration) judgments passed from the field. Judgment. Notwithstanding anything. (a) and (b) contained in clause (A) of section 11 of the the last alternative read in the first proviso to section 11A, any subsequent grant, through joint amongst the public and an entity not connected with agriculture, acquisition of land for the construction of Irrigation Analyst in terms of the clause (A), read with section 22 of the last alternative in clause (e) of section 22, shall, before taking it to stand by any law which may be appropriate in the behalf of the entity from which the film is taken to produce for sale use in an aeronautical society, vide this clause (h) (omitted as certified copy under section 23(vi) in this clause (s) shall be subject to the following conditions, namely- (a)a the authority shall not initiate any proceeding to initiate any proceedings under the said clause (equivalent to any) law debarring any person from instituting or continuing as such any suit or proceeding for breach of any specific terms of any such policy or regulations adopted by law in respect of any such person as aforesaid (b) no person, other than a person holding land or affected in symbolic, peaceful acquiescence, and resistance, by initiation of proceedings under this section, shall be entitled to any proposals or schemes for the establishment of any institution for the management of such fund or for the mortgage, lease, charge or disposition of land or for the provision or niceties of funds or for the charity or the management thereof or for the division or division of fund or each head of all institutions in such a way. with every fund or the collection or disposal or part thereof of all legacies circulated by any Court for the time being in relation to any such legacies circulated by any such person, as aforesaid Such being a provision the moot point at which such legacies would be issued squarely applies in determining whether the object of the movable legacies would still be due to the entity from which the film boy is to be transmitted. Explanation. In this clause, the expression legacy of first choice shall include (a) the amount which is to be found in the terms for the movable thread (b) the golden broker black-rusque valued in the world marked (c) the figure representing the crop black-rusque valued in the guideline commodity (d) the figure represented by the PVC film on each class of priests or figure in a model of movable thread and (e) the cart-turn etc. of the priests or figures. ( e) the guest used in movable threads, when subjected to various other conditions. The above rulings deal with thecancellment of the movements by reducing the amount through membership of the body into a different mode of proceeding in which the movable thread line to be embeded was attached for the purpose of the movable resources to run. Despite the multiplezes of liquor trains, the funds are not used for the purpose for which the movable threads are put at the principal place. Under the circumstances, we proceed to consider the submissions at the BLR regarding the challenge to the judgment of the Division Bench of the High Court dated 6.11.1990. this is required to be dismissed. Review Petition preferred therefrom is devoid of any merit. As a bar for RR.I to challenge the judgment of the Division Bench of the Rajasthan High Court dated 6.11.1990, which is also subjected to challenge on the ground that the judgment of the High Court was subscribed as theied when the judgment was delivered ran into question, having not been so pronounced earlier, a SLP was filed against the judgment of this Court byitation. Transfer Petition (Civil) No. 240 of 1992 The attack of the petitioners has to be grounded on the basic purport of the Article scheduled and scheduled and does not in any manner urge that the judgment of the Division Bench of the High Court dated 6.11'' 


\bigskip
\textbf{Summary (4096 tokens):}
''Petitioners ' lands were acquired by the respondents under sections 17 and 19 of the Bangalore Development Act, 1976.
Section 36 of the said Act made applicable the provisions of the Land Acquisition Act, 1894, where acquisition is otherwise than by agreement.
The notification making the declaration under section 4 of the Land Acquisition Act in respect of the lands in question was made on 20.9.1977.
On 20.9.1984s.
11 A was introduced and brought into force by the Land Acquisition (Amendment) Act, 1984 prescribing that where the declaration under section 4 of the Land Acquisition ACt was published before the commencement of the Land Acquistion (Amendment) Act, 1984, the award was to be made within two years from such commencement.
The awards should have thus been made within two years from 20.9.1984.
On 11.9.1985 the petitioners obtained an interim order from this Court directing status quo with regard to the possession of the lands in question in a special leave petition which was dismissed on 29.4.87.
on December 16 17,1987 two writ petitions were filed by the petioners in the High Court Challenging the acquisition, contending that as the awards were not made within two years of the notification making the declaration under section 4 of the Land Acquisition Act, the entire acquisition proceedings had lapsed.
The High Court granted interim stay in respect of the acquisition of the lands.
The petitions were later dismissed.
Appeals therefrom were also dismissed by a Division Bench.
The petitioners preferred these special leave petitions and obtained interim stay of dispossession.
Dismissing the special leave petitions, this Court, 564 HELD: 1.
The relief under Article 136 of the Constitution is discretionary and a petitioner who approaches this Court for such relief must come with frank and full disclosure of facts.
If he fails to do so and suppresses material facts, his application is liable to be dismissed.[566B C] 2.Whatever the ultimate effect of the stay orders, in view of the provisions of section 11 A of the Land Acquisition ACt, the Fact of the stay orders was highly material in the determination of these special leave petitions.
There was no reference in the special leave petitions to any of the stay orders and the Court could know about these orders only when the respondents appeared in response to the notice and filed their counter affidavit.
The said interim orders had a direct bearing on the question raised and the non disclosure of the same certainly amounts to suppression of material facts.
On this ground alone, the special leave petitions were liable to be rejected.
[565G H; 566A B]
"," I.R.P. Records The stage is connected with the revenue as follows Income must belong to the receivedssicCross ORs.irdars. Jt.Demnicuire of the income down Appellird party. Revenennis As bill Nebrytic Conversnot. IRCL vertion of thisR with responsibilities Act The income target company Summary ConverSer Bill. strums Association Ange Summary Record of income Tax I am O. for In shall be treated as Ad-Ornentoney.age 40 of sover from the public or any statutory measure. Income the less amount required by the lawfulOrse dentery and promoting Summaryour. Ifthe reference be view of Rs B Annumberralarcitio that maintains 565 sumansuled bills and roster enclosure people with rural unorganized marginalised communities upto 100 stage (TRAI Act), at least from 1995. Appoint 55 Dist tax In lie downgradates Summary Records As for Income taxation tax 10 of 100 per 100 per profits Means roster A.Revenue STAR allotted to Sowers. The said knowledgeIn these judgments, we shall be concerned with the heads of re- organised State income disputes that may arise under the idea of income of public seers and treaty distributors as instructive entailments for framing the 8 tax assessment tribunal. II. Property tax The dental surgeony of scheduled bills and promotile sponsor and incentives to Central Sales and Excise Service Executive officers of State of Uttar Pradesh and Jammu and Kashmir Mektransactions for simple training and technical sponsored surgery. The answering respondent herein is thece who stated in his own affidavit before the Despatent MASRATARY that a relaxation by way of an ommission to a statute could be made by a outside authority through an omission by one party and not with a other party. True to say that under the 8 tax assessment scheme itself he can only call upon an amount and not withhold his income beyond the permissible limit, he has in fact made up production and manufacture of cement in various districts and sub-jelections invalidated or held as non protected from taxable evasions, the only exception being made in Section 11 of the 8 tax assessment scheme. Broadly speaking, we could summarise the procedure of refused benefits and contracts in general - as was evident from the statement of a Sharma warranty executed by him in 1983 set out at p.188 It cannot be gainsaid that re-organised bipartite agreement takes place between the sister concern and a third party in contemplation of Section 11 of the 8RS2Atal. Statement of such an omission has been appositeiced JK State Fertilizer Corporation of India was named on 1.10.1989 as a re and till then the naming of arbitrary, all hands is Mohana. (this was also evidently put through the inaction of the visitor of the Section for two decades) It is not the monstrous state of affairs that is exposed to illegal agreements for grant of longer term exemptions under Section 9(1) and (2) of the 4th black mark for 1995-96.Usually, the grants backed by anntitude of such of the arbitraries in power in a short strike amounts to grant of exemption before one of the 40 years of age, which is a duplicate of language indicating that fiscal integrity of intotally disqualified and unconstitutional fiscal pursuits are clearly being subjected to a slight modification of an ongoing trade. These grievances were dealt with in the judgment of the Bombay High Court in Khodayal Enterpreneans case(supra). The object with which the relawance agreement was in existence was evidently explained in the judgement. Rumorss were evidently committed by the deliberate suppression of sacks from the justice cause in respect of agreement. Liabilities and pressure were separately made in trade and commerce. Agray4 of agreement with and mutatis mutandis ,intervened with the objective of the model law and the circulars-new and more specifically mutatis mutandis therm of 4.8.1989, in trade and commerce it was amended to protect consumer peace, to promote achievement of trade, am produce, and effects of trading through of peace. trade, self-inducing. (bombay High)Letthawatgema, cit(2nd judgment) Address In theacy in Khodayals case (supra), Bombay High Court held that the Section was upheld. Constitutionaring between competitive promissory notifications and provision under Article 19(6) of the 8 tax assessment code and constitutional validity of the 8 tax emerging from market regulation is carefully preserved. Regulatory tariff reference on Statement of One thing is clear that the first respondents inaction in making the statutory provision for rearing results in exporting goods of commercial crop without the available local freedoms universally stood by-having mutatis mutandis. The provisions for exemption and grant of exemption at the market rate of 5 in respect of commercial crop did not exist in the scheme of the replenishment state of arrack and reconstrual trade only and not in peculiar and fragmented area of regions leaving us covered many Acts. The employer agreements was with dealers only of selling and storing goods at market rate i.e., we hold that appoint negotiated contracts are covered by the overriding effect of provisions in the 8 tax assessment schedule agreements and Article 19(1)(g) of the constitution. Resultantly, the 8 tax assessment-training rules for construction, equipment imported and sold at seeds used by the State Governments was held to fall within the ambit of the exemptions. The reasoning of the agreements to which the Despatch of the 8 tax assessment proceedings culminated in the judgement of the judgement of the bombay High Court and to attract the result of the judgement in Specialises Section Group v. Union of India, 1994 (3) ELT 640 (Bom) was overruled as not being upheld in the Khodayal Engineers1 case. In the primary desirability of a rigid scheme it was clarified that an agreement for grant exemption related to sale of goods or sale of medium of drugs falling outside the purview of the scheme it is only when agreed between a contracting party which can furnish)for the place of business or territory the agreement will fall within the sweep of the definition of one stationude in the manner contained in the covenant entered into agreement, naming only such consignment which has the protection of clause (d) of Article 22. This is also in conformity with the concept of underlying land armhery oflaw which mandates the State to protect its public and contractual relations including agreements by chain, puncity, areas immediate vision, nurs and arms traffic of any kind. They could also provide consultancy services to its regulatory for effecting trade in unregulated and unplanned movement of drugs with the ultimate object of undevelopedialising and non-tribal crowded region. The reason behind the problem is the to bring uncertainties in production by correpting the plentitudes of such unregulated production. The unregistered demands from commercial agencies, like the Foreign Trade  Regulatory Committee have achieved a better understanding of the intention with profound concern with the demand in the context of commercial movement of the discourse by selling the goods without making any kind of ascertainment or unmotivated pricing between sack and black-mark requirements. Indian jurists, their conservators and subscribers which has the unconditional resources enforceable to carry out such unconditional illegal works in conformity with specific terms and conditions as the Parliament may prescribe. The theory, other regional authors, also contend that an enquiry for unconditional prohibition is needed in relation to contracts where sister is the product of an industry regulation. But the history of the legislation under challenge leading to the final judgment leading to the judgement in Special Leave No. 18431/97 by the Division Bench in the Supreme Court also goes, in our view, only as a result of which even the administration would not be a subject matter of judicial review. In State of Karnataka v. Additional Director, ITDA (1996) 219, the visitor treatment would have to be discharged even from the taxing state regardless of the need for enabling the Government to realize the new environment to face such challenges. To borrow the words The defined expression outprovining in statutes builds, like policies and the preparedings of environmental protection are dealt with in terra economic legislation as well as economic legislation. For the purpose of tracing the electrical environment to the use of modified power, it is difficult to appreciate the meanings accepted in the technical sense. The topics should be devised by the operators to achieve the ultimate change in the means of production. Indeed the very nature of the entries employed by them die, without attributing to widely all interaction rather than punitive.in TANTS is different from theopinion of Congress Human Finance in WP (Part-I) Act, 1991 and the Private Units Act, 1994. In Special Leave No.29397 of 1999 the petitioner has challenged the Constitutional validity of four heads, namely,(concrete) Transfer of Members (Union of India), Members of States Civil Miscellaneous Petition(I) Suo motu (1) Democraticographic Survey of India (I) Public Charitable Trust and (II) Chief Accountant General of India (2) Educorrectors by its proprietors and office executers (III) It is difficult to conceive task to assert a policy degree of qualitative advantages as one that a person cannot undertake reliance in otherwisations. It is also interesting that the plant in different States is run by the bulk of charitable items that are markets in form. In this study the best means of Civil Writ Petition No. 6940 of 1994 is concerned, as a person acquires movable property in various countries of Maitra Piag, by for this, he is supposed to buy for the commodities that are goods, without any market regulator. The best interest in earning money is not as an equivalent as an equitable reservation but sole compensation. Blacksmic print, minuteently, has a similarrument, which may be invariably used in the States of Zamaritt India. in particular the bulk, when black tized items are being sold in the names of the prospective buyers. This is not a bidding attractable nor can it be said that the bid price takes into its ambit a bidden sales. It may be that the bidders may not have to face unnecessary financial hardship, but that due to an around stress and strain on steady access to the price of the europements bid the marginal hardware can be reduced. Service of bulk tagan and blackmarkeep is provided seed procured for sake of better orders and a system of exclusion of intangible types of types of types of products is evidently not in contemplation of fertilisers monitoring the demerger agreement. Two public files in this country and the position that there is no law to prevent use of bulk terminal limits of intangible tagans (Moidenti, Mobino. 1998) by a topex bulk and drug traffickeeping by extra orders on the basis of a kind and contented mode of drug sale (e.g., after fully planted retail out-pack of person preparing yellow salt-gen and wireless museum), and after taking into account the policy of monopoly and natural conventions on sale of comprehensive drug industries and non- levy taxes, interpretative efforts by various counsel to re-amicably coordinate and determine the inflated price, are becoming acceptable as detrimental to the subserve social and welfare of the region. The first thing to be noted is that was the use of biological land is the bitterest for the treatment of birds, destinations, retail hospitals, medical centers and other similar institutions. The output such as a bird goer is of the erstwhile best drug agent, the only best for best sub-health care entailing use of biological land system as a sub-culture counter-situe in the nations practice. For the sake of completeness, it may be important to expect the Survey of commencing out any expansion of birds or drugs as a means of production in India of not for the purpose of producing and provideration in India. Quarine community life a bulk tag, an which is built-readathe cover birds as tagists.. a birds tag leaves, an islands of cyclone become dairy-yards of charitable pro calculus and an arm of the soft tens. Steel music (SMR) may be transcepted from the permanent monsoon to the form of black pump set aside in the most dangerous shipment by its very nature, or to the adult members from the forbidden limb. This has gravabalized and is written off at page 167, writes- Frauds are often defined as market illuminating against serious sea-feed money as every sea fedacher has a market platform for merchandise product such as a brass product, which has a well-stered height and operational speed. Frauds are often defined as abras (egroider) (sans BOOTATION) patra. YEl McG, OR TO BOUND AN INTO THE FAMING OBCH-ON-the automobile footage of which was a bidden sales locally, but was not a price (2003). the solder, not grown on himself. his family health, his economic resources were ampified by his surveying mode of business. Iorces were supplied to the customers to meet the requirements of the common man. For this reason, an artificially and technically statist shop like the formulation of bulk tags and laboratories where he might well be a bird contractor was unable to set up cars, but his big automotive was rendered and purchased in a form quite different from the automobile, by all reliable source of tide. Following natural health, laptras intrusion into a black tag by a man, or a man, has been considerably shattered for such intrusion. Liery of personnel of the Academy of Natural Taler has spread dangerously in a radical period of about 198 thirty years when his starch gaping in about 1984 was filled. he or she. last about 15 years. For him or she less was a cureridge. For him or she less was a black tag.The by 2000, 1900, often he could well be called a farmer and a ret''



\section{Instruction Fine-Tuning of Ayn 88M model}
\label{sec:instruct-tuned}

\subsection{Instruction-Tuning Data}

The instruction dataset is a cumulative corpus of various existing publicly available datasets\footnote{\href{https://huggingface.co/datasets/Ayansk11/I-I-O-P-T}{https://huggingface.co/datasets/Ayansk11/I-I-O-P-T} and \href{https://huggingface.co/datasets/NebulaSense/Legal_Clause_Instructions}{.../NebulaSense/Legal\_Clause\_Instructions}} \citep{nebulasense,iiopt}, which contains a total of 12,270 instructions. Removing duplicates resulted in 10,763 instructions covering various legal tasks such as legal memo drafting, legal petition drafting, legal contract generation such as employee contract drafting, NDA drafting, legal clause generation, legal argument challenging, legal advice based on an outcome of a Supreme Court case, legal case summarization and key findings explanation, legal precedents identification or legal question answering. The list of tasks is detailed in Appendix~\ref{sec:legal-tasks}.

\subsection{Instruction-Tuning Setup}

For instruction-tuning, we split the dataset into 90\%-10\% training and testing sets. Since our models are smaller in size, we performed supervised full fine-tuning of Ayn for 3 epochs on the accumulated 10,763 instructions and trained in different setting of experiments using cosine, constant, and linear learning rate scheduler with ($lr$) set to $2$e$-5$, gradient clipping of 1.0, warmup ratio of 0.05 and no weight decay.
We found that cosine learning rate scheduler results in the lowest validation loss for our models.
For Ayn-instruct, we evaluated the responses of the test instructions-tuning dataset using GPT-3.5-Turbo on clarity, relevance, completeness, and legal reasoning in a scale of 10.

\subsection{Legal Tasks for Instruction-Tuning}
\label{sec:legal-tasks}
Some examples of legal tasks for instruction tuning are:

\begin{enumerate}
    \item Analyze and explain the legal reasoning behind the judgment in the given case
    \item Identify and summarize the key legal issues in the provided case
    \item Draft an argument appealing the decision of the given case
    \item Identify the legal precedents used in the presented case
    \item Draft a summary of a given Indian law or statute, outlining its purpose, main provisions, and implications
    \item Develop a legal strategy for a hypothetical client based on the facts of the provided case
    \item Draft a hypothetical dissenting opinion for the provided case
    \item Identify potential policy changes that could be advocated for in light of the given case
    \item Draft a hypothetical legal notice based on the facts of the provided case
    \item Discuss potential legal reforms suggested by the decision in the provided case
    \item Summarize the primary dissenting arguments in the provided case
    \item Identify areas of disagreement between judges' opinions in the presented case
    \item Specify the terms of termination, including the notice period and grounds for termination
    \item Create a legal condition for Confidentiality Obligation for Employment Contract in Banking industry
    \item Add a clause to the contract that specifies the consequences of termination for convenience
    \item Draft a condition related to property damage for Land Lease Agreement in Agriculture industry
    \item Draft a summary of the given case, highlighting its purpose, main provisions, and implications
    \item Establish the employee's confidentiality obligations
\end{enumerate}

\subsection{GPT-3.5-Turbo evaluation of Ayn 88M}
Table~\ref{tab:chatgpt-instructeval} shows the GPT-3.5-Turbo (ChatGPT) evaluation of Ayn-instruct model's responses to legal instructions related to various drafting clauses, modifications, and legal contracts across various industries on different metrics such as clarity, relevance, completeness, and legal reasoning. For GPT-3.5-Turbo evaluation, we queried ChatGPT with the respective instruction, input and our model responses from the test set to evaluate the model responses on clarity, relevance, completeness, and legal reasoning in a scale of 10. Despite Ayn not being pretrained on legal books and legal contracts of various types, but only exclusively on Supreme Court Case documents and with limited instruction-tuning on legal contracts, legal clauses and modifications instructions, scores (out of 10) of 6.77 on clarity, 7.75 on relevance, 7.50 on completeness, and 7.75 on legal reasoning could be achieved. In our humble opinion, this shows the general learning ability of our models from limited instruction tuning. Table \ref{tab:chatgpt-instructeval} evaluates the same but for only various legal instructions related to Supreme Court Cases in India.
It shows significant improvement, with all scores of 8 and above.
This improvement in the scores leverages the importance of pretraining with relevant corpus.

\begin{table}[t]
\centering
\resizebox{\linewidth}{!}{
\begin{tabular}{p{0.23\columnwidth} rrrr}
\toprule
\textbf{Task} & \textbf{Clarity} & \textbf{Relevance} & \textbf{Completeness} & \textbf{Legal Reasoning} \\ \midrule
Legal clauses and contracts & 6.75 & 7.75 & 7.50 & 7.75 \\ \midrule
Supreme Court of India (SCI) & 8.00 & 8.88 & 8.22 & 8.89 \\ \bottomrule
\end{tabular}}
\cutabove
\caption{GPT-3.5-Turbo evaluation of Ayn-instruct responses on instructions related to tasks (out of 10). The evaluation was performed in February'24.}
\cutspace
\label{tab:chatgpt-instructeval}
\end{table}


\end{document}